\documentclass{article}
\usepackage{spconf,amsmath,graphicx}
\usepackage{bm}
\usepackage[subrefformat=parens]{subcaption}
\usepackage{multirow}
\usepackage{amssymb}
\usepackage{cite}
\usepackage{threeparttable}
\usepackage{cite}
\newcommand{\argmax}{\mathop{\rm argmax}\limits}

\usepackage{multirow}
\usepackage{amsfonts}

\title{Hierarchical transformer-based Large-context End-to-end ASR \\ with Large-context Knowledge Distillation}
\name{Ryo Masumura, Naoki Makishima, Mana Ihori, Akihiko Takashima, Tomohiro Tanaka, Shota Orihashi}
\address{NTT Media Intelligence Laboratories, NTT Corporation, Japan}
\begin{document}
\ninept
\maketitle
\begin{abstract}
  We present a novel large-context end-to-end automatic speech recognition (E2E-ASR) model and its effective training method based on knowledge distillation. Common E2E-ASR models have mainly focused on utterance-level processing in which each utterance is independently transcribed. On the other hand, large-context E2E-ASR models, which take into account long-range sequential contexts beyond utterance boundaries, well handle a sequence of utterances such as discourses and conversations. However, the transformer architecture, which has recently achieved state-of-the-art ASR performance among utterance-level ASR systems, has not yet been introduced into the large-context ASR systems. We can expect that the transformer architecture can be leveraged for effectively capturing not only input speech contexts but also long-range sequential contexts beyond utterance boundaries. Therefore, this paper proposes a hierarchical transformer-based large-context E2E-ASR model that combines the transformer architecture with hierarchical encoder-decoder based large-context modeling. In addition, in order to enable the proposed model to use long-range sequential contexts, we also propose a large-context knowledge distillation that distills the knowledge from a pre-trained large-context language model in the training phase. We evaluate the effectiveness of the proposed model and proposed training method on Japanese discourse ASR tasks.
\end{abstract}
\section{Introduction}
In the automatic speech recognition (ASR) field, end-to-end ASR (E2E-ASR) systems that directly model a transformation from an input speech to a text have attracted much attention. Main advantage of E2E-ASR systems is that they allow a single model to perform the transformation process in which multiple models, i.e., an acoustic model, language model, and pronunciation model, have to be used in classical ASR systems. In fact, E2E-ASR systems have the potential to perform overall optimization for not only utterance-level processing but also discourse-level or conversation-level processing.

Toward improving E2E-ASR performance, several modeling methods have been developed in the last few years. The initial studies mainly adopted connectionist temporal classification \cite{zweig_icassp2017,audhkhasi_interspeech2017} and recurrent neural network (RNN) encoder-decoders \cite{bahdanau_icassp2015,lu_interspeech2015}. Recent studies have used the transformer encoder-decoder, which provided much stronger ASR performance \cite{vaswani_nips2017,dong_icassp2018}. The key strength of the transformer is that relationships between the input speech and output text can be effectively captured using a multi-head self-attention mechanism and multi-head source-target attention mechanism.

These modeling methods have mainly focused on utterance-level ASR in which each utterance is independently transcribed. Unfortunately, utterance-level ASR models cannot capture the relationships between utterances even when transcribing a series of utterances such as discourse speech and conversation speech. On the other hand, large-context E2E-ASR models, which take into account long-range sequential contexts beyond utterance boundaries, have received increasing attention. Previous studies reported that large-context models outperform utterance-level models in discourse or conversation ASR tasks \cite{kim_slt2018,masumura_icassp2019}, and hierarchical RNN encoder-decoder modeling has been mainly introduced into the large-context E2E-ASR models. However, the transformer architecture has not yet been introduced into the large-context ASR systems. The transformer architecture is expected to be leveraged for effectively capturing not only input speech contexts but also long-range sequential contexts beyond utterance boundaries. 

In this paper, we propose a hierarchical transformer-based large-context E2E-ASR model that combines the transformer architecture with hierarchical encoder-decoder based large-context modeling. The key advantage of the proposed model is that a hierarchical transformer-based text encoder, which is composed of token-level transformer encoder blocks and utterance-level transformer encoder blocks, is used to convert all preceding sequential contexts into continuous representations. In the decoder, both the continuous representations produced by the hierarchical transformer and input speech contexts are simultaneously taken into consideration using two multi-head source-target attention layers. Moreover, since it is difficult to effectively exploit the large-contexts beyond utterance boundaries, we also propose a large-context knowledge distillation method using a large-context language model \cite{lin_emnlp2015,wang_acl2016,masumura_interspeech2018,masumura_interspeech2019,masumura_asru2019}. This method enables our large-context E2E-ASR model to use the large-contexts beyond utterance boundaries by mimicking the behavior of the pre-trained large-context language model. In experiments on discourse ASR tasks using a corpus of spontaneous Japanese, we demonstrate that the proposed model provides ASR performance improvements compared with conventional transformer-based E2E-ASR models and conventional large-context E2E-ASR models. We also show that our large-context E2E-ASR model can be effectively constructed using the proposed large-context knowledge distillation. 

\section{Related Work}

\smallskip
\noindent {\bf Large-context encoder-decoder models: } Large-context encoder-decoder models that can capture long-range linguistic contexts beyond sentence boundaries or utterance boundaries have received significant attention in E2E-ASR \cite{kim_slt2018,masumura_icassp2019}, machine translation \cite{wang_emnlp2018,maruf_acl2018}, and some natural language generation tasks \cite{serban_aaai2016,ihori_icassp2020}. In recent studies, transformer-based large-context encoder-decoder models have been introduced in machine translation \cite{zhang_emnlp2018,tan_emnlp2019}. In addition, a fully transformer-based hierarchcal architecture similar to our transformer architecture was recently proposed in a document summarization task \cite{liu_acl2019}. To the best of our knowledge, this paper is the first study that introduces the hierarchical transformer architecture into large-context E2E-ASR modeling. 

\smallskip
\noindent {\bf Knowledge distillation for E2E-ASR: } For E2E-ASR modeling, various knowledge distillation methods have been proposed. The main objective is to construct compact E2E-ASR models by distilling the knowledge from computationally rich models \cite{huang_is2018,kim_icassp2019,munim_icassp2019}. Methods for distilling the knowledge from models other than ASR models into E2E-ASR models have also been examined recently. Representative methods are used to distill knowledge from an external language model to improve the capturing of linguistic contexts \cite{bai_interspeech2019,futami_arxiv2020}. Our proposed large-context knowledge distillation method is regarded as an extension of the latter methods to enable the capturing of all preceding linguistic contexts beyond utterance boundaries using large-context language models \cite{lin_emnlp2015,wang_acl2016,masumura_interspeech2018,masumura_interspeech2019,masumura_asru2019}.

\section{Hierarchical transformer-based Large-context E2E-ASR Model}
This section details our hierarchical transformer-based large-context E2E-ASR model that integrates the transformer encoder-decoder with hierarchical encoder-decoder modeling. Large-context end-to-end ASR can effectively handle a series of utterances, i.e., conversation-level data or discourse-level data, while utterance-level end-to-end ASR handles each utterance independently.

In our hierarchical transformer-based large-context E2E-ASR model, the generation probability of a sequence of utterance-level texts ${\cal W}=\{\bm{W}_1,\cdots,\bm{W}_T\}$ is estimated from a sequence of utterance-level speech ${\cal X}=\{\bm{X}_1,\cdots,\bm{X}_T\}$, where $\bm{W}_t=\{w_{t,1},\cdots,w_{t,N_t}\}$ is the $t$-th utterance-level text composed of tokens and $\bm{X}_t=\{\bm{x}_{t,1}, \cdots, \bm{x}_{t,M_t}\}$ is the $t$-th utterance-level speech composed of acoustic features. The notation $T$ is the number of utterances in a series of utterances, $N_t$ is the number of tokens in the $t$-th text, and $M_t$ is the number of acoustic features in the $t$-th utterance. The generation probability of $\cal W$ is defined as
\begin{multline}
   \begin{split}
     P({\cal W}|{\cal X},\bm{\Theta}) & = \prod_{t=1}^{T} P(\bm{W}_{t}|\bm{W}_{1:t-1},\bm{X}_{t},\bm{\Theta})  \\
     & = \prod_{t=1}^{T} \prod_{n=1}^{N_t} P(w_{t,n}|w_{t,1:n-1}, \bm{W}_{1:t-1},\bm{X}_{t},\bm{\Theta}) , \\
   \end{split}
\end{multline}
where $\bm{\Theta}$ is the model parameter set. Utterances from the 1-st to $t-1$-th utterance is defined as $\bm{W}_{1:t-1} = \{\bm{W}_1,\cdots,\bm{W}_{t-1}\}$, and tokens from the 1-st to $n-1$-th token for the $t$-th utterance is defined as $w_{t,1:n-1} = \{w_{t,1},\cdots, w_{t,n-1}\}$. 

ASR decoding of a sequence of utterance-level texts from a sequence of utterance-level acoustic features using large-context end-to-end ASR is achieved by recursively conducting utterance-level decoding. The ASR decoding problem for the $t$-th utterance is defined as
\begin{equation}
  \hat{\bm{W}}_t = \argmax_{\bm{W}_t} P(\bm{W}_t|\hat{\bm{W}}_{1:t-1},\bm{X}_t,\bm{\Theta}) ,
\end{equation}
where $\hat{\bm{W}}_{1:t-1}$ are ASR outputs from the $1$-st utterance to the $t-1$-th utterance. Therefore, $\hat{\bm{W}}_t$ is recursively used for decoding the text of the $t+1$-th utterance.

\begin{figure}[t]
  \begin{center}
    \includegraphics[width=85mm]{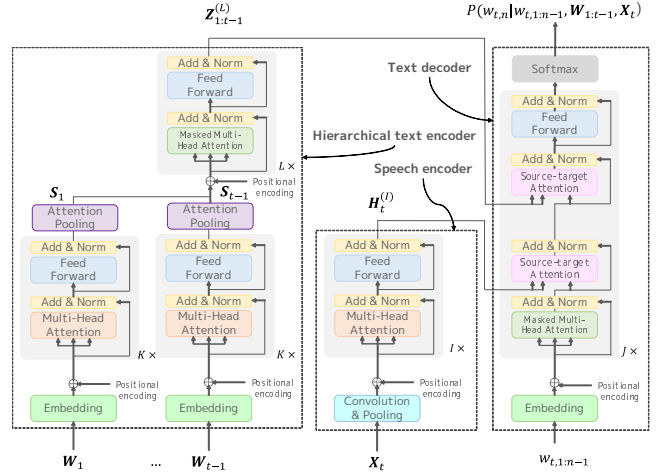}
  \end{center}
  \vspace{-4mm}
  \caption{Network structure of our hierarchical transformer based large-context E2E-ASR model.}
\end{figure}

\subsection{Network structure}
We construct our hierarchical transformer-based E2E-ASR model using a hierarchical text encoder, speech encoder, and text decoder. Thus, we define the model parameter set as $\bm{\Theta} = \{\bm{\theta}_{\tt henc}, \bm{\theta}_{\tt senc}, \bm{\theta}_{\tt dec}\}$ where $\bm{\theta}_{\tt henc}$, $\bm{\theta}_{\tt senc}$ and $\bm{\theta}_{\tt dec}$ are the parameters of the hierarchical text encoder, the speech encoder and the text decoder, respectively. Figure 1 shows the network structure of the proposed model. Each component is detailed as follows.

\smallskip
\noindent {\bf Hierarchical text encoder: } The hierarchical text encoder, which is constructed from token-level transformer blocks and utterance-level transformer blocks, embeds all preceding pre-decoded text into continuous vectors. In fact, to perform ASR for the $t$-th utterance, we can efficiently produce the continuous vectors by only feeding the $t-1$-th pre-decoded text.

For the $t-1$-th pre-decoded text, the $n$-th token is first converted into a continuous vector $\bm{c}_{t-1, n}^{(0)} \in \mathbb{R}^{d \times 1}$ by 
\begin{equation}
 \bm{w}_{t-1, n} = {\tt Embedding}(w_{t-1, n}; \bm{\theta}_{\tt henc}) ,
\end{equation}
\begin{equation}
 \bm{c}_{t-1, n}^{(0)} = {\tt AddPosEnc}(\bm{w}_{t-1, n}) ,
\end{equation}
where ${\tt Embedding}()$ is a function to convert a token into a continuous vector and ${\tt AddPosEnc}()$ is a function that adds a continuous vector in which position information is embedded. The notation $d$ represents the hidden representation size. The continuous vectors $\bm{C}^{(0)}_{t-1} = \{\bm{c}_{t-1, 1}^{(0)},\cdots,\bm{c}_{t-1, N_t}^{(0)}\} \in \mathbb{R}^{d \times N_t}$ are embedded into an utterance-level continuous vector using $K$ transformer encoder blocks and an attention pooling layer. The $k$-th transformer encoder block creates the $k$-th hidden representations $\bm{C}^{(k)}_{t-1} \in \mathbb{R}^{d \times N_t}$ from the lower layer inputs $\bm{C}^{(k-1)}_{t-1}$. The $t-1$-th utterance-level continuous vector $\bm{S}_{t-1} \in \mathbb{R}^{d \times 1}$ is produced by
\begin{equation}
 \bm{C}^{(k)}_{t-1} = {\tt TransformerEnc}(\bm{C}^{(k-1)}_{t-1}; \bm{\theta}_{\tt henc}) ,
\end{equation}
\begin{equation}
 \bm{S}_{t-1} = {\tt AttentionPooling}(\bm{C}_{t-1}^{(K)}; \bm{\theta}_{\tt henc}) ,
\end{equation}
where ${\tt TransformerEnc}()$ is a function of the transformer encoder block that consists of a multi-head self-attention layer and position-wise feed-forward network \cite{vaswani_nips2017}. ${\tt AttentionPooling}()$ is a function that uses an attention mechanism to summarize several continuous vectors as one continuous vector \cite{lin_iclr2017}.

Next, to take utterance-level sequential contexts into consideration, we embed utterance-level position information as
\begin{equation}
 \bm{Z}^{(0)}_{t-1} = {\tt AddPosEnc}(\bm{S}_{t-1}) . 
\end{equation}
We then produce the $t-1$-th context-dependent utterance-level continuous vector from $\bm{Z}^{(0)}_{1:t-1} = \{\bm{Z}^{(0)}_{1},\cdots,\bm{Z}^{(0)}_{t-1}\} \in \mathbb{R}^{d \times (t-1)}$ using $L$ utterance-level masked transformer encoder blocks. The $t-1$-th context-dependent utterance-level continuous vector $\bm{Z}_{t-1}^{(L)} \in \mathbb{R}^{d \times 1}$ is computed from
\begin{equation}
 \bm{Z}_{t-1}^{(l)} = {\tt MaskedTransformerEnc}(\bm{Z}_{1:t-1}^{(l-1)}; \bm{\theta}_{\tt henc}) ,
\end{equation}
where ${\tt MaskedTransformerEnc}()$ is a masked transformer encoder block that consists of a masked multi-head self-attention layer and position-wise feed-forward network. Finally, we construct vectors $\bm{Z}_{1:t-1}^{(L)} \in \mathbb{R}^{d \times (t-1)}$ by concatenating $\bm{Z}_{t-1}^{(L)}$ with pre-computed ones $\bm{Z}_{1:t-2}^{(L)} \in \mathbb{R}^{d \times (t-2)}$. 

\smallskip
\noindent {\bf Speech encoder:} The speech encoder converts input acoustic features into continuous representations, which are used in the text decoder. For transcribing the $t$-th utterance's input speech, the speech encoder first converts the acoustic features $\bm{X}_t \in \mathbb{R}^{f \times M_t}$ into subsampled representations $\bm{H}^{(0)}_t = \{\bm{h}^{(0)}_{t,1}, \cdots, \bm{h}^{(0)}_{t,M_t^\prime}\} \in \mathbb{R}^{d \times M_t^\prime}$ as
\begin{equation}
 \bm{H}_{t} = {\tt ConvolutionPooling}(\bm{X}_t; \bm{\theta}_{\tt senc}) ,
\end{equation}
\begin{equation}
 \bm{h}^{(0)}_{t,m} = {\tt AddPosEnc}(\bm{h}_{t,m}) , 
\end{equation}
where ${\tt ConvolutionPooling}()$ is a function composed of convolution layers and pooling layers. The notation $M_t^{\prime}$ is the subsampled sequence length of the $t$-th input speech depending on the function. Next, the speech encoder converts the hidden representations $\bm{H}^{(0)}_t$ into $\bm{H}_t^{(I)} \in \mathbb{R}^{d \times M_t^\prime}$ using $I$ transformer encoder blocks. The $i$-th transformer encoder block creates the $i$-th hidden representations $\bm{H}_t^{(i)} \in \mathbb{R}^{d \times M_t^\prime}$ from the lower layer inputs $\bm{H}_t^{(i-1)}$ by 
\begin{equation}
 \bm{H}_t^{(i)} = {\tt TransformerEnc}(\bm{H}_t^{(i-1)}; \bm{\theta}_{\tt senc}) .
\end{equation}

\smallskip
\noindent {\bf Text decoder:} The text decoder computes the generation probability of a token from the hidden representations of the speech and all preceding tokens of not only the target utterance but also preceding utterances.

We detail the procedure of estimating the generation probability of the $n$-th token for the $t$-th utterance. We first convert pre-estimated tokens $w_{t,1:t-1}$ for the $t$-th output text into continuous vectors $\bm{u}_{t,1:n-1}^{(0)} \in \mathbb{R}^{d \times (n-1)}$ as 
\begin{equation}
 \bm{w}_{t,1:n-1} = {\tt Embedding}(w_{t,1:n-1}; \bm{\theta}_{\tt dec}) , 
\end{equation}
\begin{equation}
 \bm{u}_{t,1:n-1}^{(0)} = {\tt AddPosEnc}(\bm{w}_{t,1:n-1}) .
\end{equation}
Next, the text decoder integrates $\bm{u}_{t,1:n-1}^{(0)}$ with input speech contexts $\bm{H}_t^{(I)}$ and all preceding linguistic contexts $\bm{Z}_{1:t-1}^{(L)}$ using $J$ transformer decoder blocks. The $j$-th transformer decoder block creates the $j$-th hidden representation $\bm{u}_{t,n-1}^{(j)} \in \mathbb{R}^{d \times 1}$ from the lower layer inputs $\bm{u}_{1:n-1}^{(j-1)} \in \mathbb{R}^{d \times n-1}$ by
\begin{equation}
 \bm{u}_{t, n-1}^{(j)} = {\tt TransformerDec}(\bm{U}^{(j-1)}_{t, 1:n-1}, \bm{H}_t^{(I)}, \bm{Z}_{1:t-1}^{(L)}; \bm{\theta}_{\tt dec}) ,
\end{equation}
where ${\tt TransformerDec}()$ is a transformer decoder block that consists of a masked multi-head self-attention layer, two multi-head source-target attention layers, and a position-wise feed-forward network. In the multi-head source-target attention layers, we first use $\bm{H}_t^{(I)}$ then use $\bm{Z}_{1:t-1}^{(L)}$ as the source information. The predicted probabilities of the $n$-th token for the $t$-th utterance $w_{t,n}$ are calculated as
\begin{equation}
 P(w_{t,n}|w_{t,1:n-1}, \bm{W}_{1:t-1},\bm{X}_{t},\bm{\Theta}) = {\tt Softmax}(\bm{u}_{t,n-1}^{(J)}; \bm{\theta}_{\tt dec}) ,
\end{equation}
where ${\tt Softmax}()$ is a softmax layer with a linear transformation. 

\subsection{Training}
The model parameter sets can be optimized from training datasets ${\cal D} = \{({\cal X}^1, {\cal W}^1),\cdots, ({\cal X}^{|\cal D|}, {\cal W}^{|\cal D|})\}$, where $|\cal D|$ is the number of conversation-level or discourse-level data elements in the training datasets. The $a$-th data element is represented as ${\cal X}^a=\{\bm{X}^{a}_1,\cdots,\bm{X}_{T^a}^a\}$ and ${\cal W}^a=\{\bm{W}^{a}_1,$ $\cdots,$ $\bm{W}_{T^a}^{a}\}$, where $\bm{W}^{a}_t=\{w_{t,1}^{a},$ $\cdots,$ $w_{t,N_t^{a}}^{a}\}$. The loss function to optimize the model parameter sets with the maximum likelihood criterion is defined as 
\begin{multline}
 {\cal L}(\bm{\Theta}) = - \sum_{a=1}^{|\cal D|} \sum_{t = 1}^{T^a} \sum_{n=1}^{N_t^a} \sum_{w_{t,n}^a \in{\cal V}} {\hat P}(w_{t,n}^a|w_{t,1:n-1}^a,\bm{W}_{1:t-1}^a, \bm{X}_{t}^a) \\
 \log P(w_{t,n}^a|w_{t,1:n-1}^a,\bm{W}_{1:t-1}^a, \bm{X}_{t}^a, \bm{\Theta}) ,
\end{multline}
where $\cal V$ is the vocabulary set. ${\hat P}$ represents the ground-truth probability that is 1 when $w_{t,n}^a = \hat{w}_{t,n}^a$, and 0 when $w_{t,n}^a \neq \hat{w}_{t,n}^a$. Note that $\hat{w}_{t,n}^a$ is the $n$-th reference token in the $t$-th utterance in the $a$-th element.

\section{Large-Context Knowledge Distillation}
This section details our proposed large-context knowledge distillation method as an effective training method of large-context E2E-ASR models. Our key idea is to mimic the behavior of a large-context language model \cite{lin_emnlp2015,wang_acl2016,masumura_interspeech2018,masumura_interspeech2019,masumura_asru2019} pre-trained from the same training datasets. A large-context language model defines the generation probability of a sequence of utterance-level texts ${\cal W}=\{\bm{W}_1,\cdots,\bm{W}_T\}$ as
\begin{equation}
 P({\cal W}|\bm{\Lambda}) = \prod_{t=1}^{T} \prod_{n=1}^{N^t} P(w_{t,n}|w_{t,1:n-1}, \bm{W}_{1:t-1},\bm{\Lambda}) , \\
\end{equation}
where $\bm{\Lambda}$ is the model parameter set for the model. For the network structure, we use the hierarchical text encoder and the text decoder described in Section 3.1. A loss function to optimize $\bm{\Lambda}$ is defined as 
\begin{multline}
 {\cal L}(\bm{\Lambda}) = - \sum_{a=1}^{|\cal D|} \sum_{t = 1}^{T^a} \sum_{n=1}^{N_t^a} \sum_{w_{t,n}^a \in{\cal V}} {\hat P}(w_{t,n}^a|w_{t,1:n-1}^a,\bm{W}_{1:t-1}^a) \\
 \log P(w_{t,n}^a|w_{t,1:n-1}^a,\bm{W}_{1:t-1}^a, \bm{\Lambda}) .
\end{multline}

We use the pre-trained parameter $\hat{\bm{\Lambda}}$ for a target smoothing of the large-context E2E-ASR training.
With our proposed large-context knowledge distillation method, a loss function to optimize $\bm{\Theta}$ is defined as
\begin{multline}
 {\cal L}_{\tt kd}(\bm{\Theta}) = - \sum_{a=1}^{|\cal D|} \sum_{t = 1}^{T^a} \sum_{n=1}^{N_t^a} \sum_{w_{t,n}^a \in{\cal V}} {\tilde P}(w_{t,n}^a|w_{t,1:n-1}^a,\bm{W}_{1:t-1}^a, \bm{X}_{t}^a) \\
 \log P(w_{t,n}^a|w_{t,1:n-1}^a,\bm{W}_{1:t-1}^a, \bm{X}_{t}^a, \bm{\Theta}) ,
\end{multline}
\begin{multline}
 {\tilde P}(w_{t,n}|w_{t,1:n-1},\bm{W}_{1:t-1}, \bm{X}_{t}) = \\
 (1-\alpha) {\hat P}(w_{t,n}|w_{t,1:n-1},\bm{W}_{1:t-1}, \bm{X}_{t}) + \\
 \alpha P(w_{t,n}|w_{t,1:n-1},\bm{W}_{1:t-1} , \hat{\bm{\Lambda}}) ,
\end{multline}
where $\alpha$ is a smoothing weight to adjust the smoothing term. Thus, target distributions are smoothed by distributions computed from the pre-trained large-context language model. Note that this target smoothing is regarded as an extension of label smoothing \cite{szegedy_cvpr2016} to enable the use of contexts beyond utterance boundaries. 

\section{Experiments}
The effectiveness of the proposed model and method were evaluated on Japanese discourse ASR tasks using the Corpus of Spontaneous Japanese (CSJ) \cite{maekawa_lrec2000}. We divided the CSJ into a training set (Train), validation set (Valid), and three test sets (Test 1, 2, and 3). The validation set was used for optimizing several hyper parameters. The segmentation of each discourse-level speech into utterances followed a previous study \cite{hori_interspeech2017}. We used characters as the tokens. Details of the datasets are given in Table 1.

\begin{table}[t!]
  \caption{Experimental datasets}
  \vspace{-3mm}
 \label{}
 \scriptsize
 \begin{center}
  \begin{tabular}{|l|rrrr|} \hline
    & Data size & Number of & Number of & Number of  \\
    & (Hours) & lectures & utterances & characters  \\ \hline \hline
    Train & 512.6 & 3,181 & 413,240 & 13,349,780 \\
    Valid & 4.8 & 33 & 4,166 & 122,097 \\
    Test 1 & 1.8 & 10 & 1,272 & 48,064  \\
    Test 2 & 1.9 & 10 & 1,292 & 47,970  \\ 
    Test 3 & 1.3 & 10 & 1,385 & 32,089  \\ \hline
  \end{tabular}
 \end{center}
 \vspace{-8mm}
\end{table}

\subsection{Setups}
We compared our proposed hierarchical transformer-based large-context E2E-ASR model with an RNN-based utterance-level E2E-ASR model \cite{bahdanau_icassp2015}, transformer-based utterance-level E2E-ASR model \cite{dong_icassp2018}, and hierarchical RNN-based large-context E2E-ASR model \cite{masumura_icassp2019}. We used 40 log mel-scale filterbank coefficients appended with delta and acceleration coefficients as acoustic features. The frame shift was 10 ms. The acoustic features passed two convolution and max pooling layers with a stride of 2, so we down-sampled them to $1/4$ along with the time axis. For the RNN-based models, the same setup as in previous studies were used \cite{masumura_icassp2019}. For the hierarchical text encoder, we stacked two token-level transformer encoder blocks and two utterance-level transformer encoder blocks. For the speech encoder and text decoder, we stacked eight transformer encoder blocks and six transformer decoder blocks. The transformer blocks were created under the following conditions: the dimensions of the output continuous representations were set to 256, dimensions of the inner outputs in the position-wise feed forward networks were set to 2,048, and number of heads in the multi-head attentions was set to 4. In the nonlinear transformational functions, the GELU activation was used. The output unit size, which corresponds to the number of characters in the training set, was set to 3,084. We also constructed a hierarchical transformer-based large-context language model. The network structure was almost that same as our proposed hierarchical transformer-based large-context E2E-ASR model.

For the mini-batch training, we truncated each lecture to 50 utterances. The mini-batch size was set to 4, and the dropout rate in the transformer blocks was set to 0.1. We used the Radam \cite{liu_iclr2020} for optimization. The training steps were stopped based on early stopping using the validation set. We also applied SpecAugment with frequency masking and time masking \cite{park_is2019}, where the number of frequency masks and time-step masks were set to 2, frequency-masking width was randomly chosen from 0 to 20 frequency bins, and time-masking width was randomly chosen from 0 to 100 frames. We also applied label smoothing \cite{szegedy_cvpr2016} and knowledge distillation using a pre-trained language model \cite{bai_interspeech2019} to the utterance-level E2E-ASR models, and applied label smoothing and our proposed large-context knowledge distillation method using a pre-trained large-context language model to the large-context E2E-ASR models. Hyper-parameters were tuned using the validation set. For ASR decoding using both the utterance-level and large-context end-to-end ASR, we used a beam search algorithm in which the beam size was set to 4.

\subsection{Results}
Tables 2--4 show the evaluation results in terms of character error rate (\%).Table 2 shows the results of comparing our proposed hierarchical transformer-based large-context E2E-ASR model with the above-mentioned conventional models (we did not introduce target smoothing into each model). The results indicate that the proposed model improved ASR performance compared with the transformer-based utterance-level model and RNN-based large-context model. This indicates that the large-context architecture of the proposed model can effectively capture long-range sequential contexts while retaining the strengths of the transformer. Table 3 shows the results of using oracle preceding contexts for the proposed model to reveal whether recognition errors of the preceding contexts affect ASR performance. The results indicate that using oracle contexts is comparable with using ASR hypotheses. This indicates that recognition errors in the preceding contexts rarely affect total ASR performance. Table 4 shows the results of applying label smoothing, knowledge distillation (KD in this table) and our proposed large-context knowledge distillation (large-context KD in this table) to the E2E-ASR models for target smoothing. The results indicate that our proposed large-context knowledge distillation effectively improved our hierarchical transformer-based large-context E2E-ASR model compared with no target smoothing and label smoothing. This confirms that our large-context knowledge distillation, which mimics the behavior of a pre-trained large-context language model, enables a large-context E2E-ASR model to use large-contexts. These results indicate that our proposed hierarchical transformer-based large-context model with our large-context knowledge distillation method is effective in discourse-level ASR tasks.

\begin{table}[t!]
  \label{}
  \scriptsize
  \begin{center}
    \caption{Comparison with conventional models} 
    \vspace{-3mm} 
    \begin{tabular}{|ll|rrr|} \hline 
      Model & ASR system & Test 1 & Test 2 & Test 3 \\  \hline \hline
      RNN \cite{bahdanau_icassp2015} & Utterance-level & 8.9 & 6.7 & 7.9 \\
      Transformer \cite{dong_icassp2018} & Utterance-level & 7.6 & 5.9 & 6.0 \\
      Hierarchical RNN \cite{masumura_icassp2019} & Large-context & 8.4 & 6.2 & 7.2 \\
      Hierarchical transformer & Large-context & {\bf 7.0} & {\bf 5.3} & {\bf 5.5} \\ \hline
    \end{tabular}
  \end{center}
  \vspace{-8mm}
\end{table}

\begin{table}[t!]
  \label{}
  \scriptsize
  \begin{center}
    \caption{Effect of ASR errors in preceding contexts} 
    \vspace{-3mm} 
    \begin{tabular}{|ll|rrr|} \hline 
      Model & Preceding contexts & Test 1 & Test 2 & Test 3 \\  \hline \hline
      Transformer \cite{dong_icassp2018} & - & 7.6 & 5.9 & 6.0 \\
      Hierarchical transformer & Hypotheses &  {\bf 7.0} & {\bf 5.3} & 5.5 \\ 
      Hierarchical transformer & Oracle &  {\bf 7.0} & {\bf 5.3} & {\bf 5.4} \\ \hline
    \end{tabular}
  \end{center}
  \vspace{-8mm}
\end{table}

\begin{table}[t!]
  \label{}
  \scriptsize
  \begin{center}
    \caption{Effect of large-context knowledge distillation} 
    \vspace{-3mm} 
    \begin{tabular}{|ll|rrr|} \hline 
      Model & Target smoothing & Test 1 & Test 2 & Test 3 \\ \hline \hline
      Transformer \cite{dong_icassp2018} & - & 7.6 & 5.9 & 6.0 \\
      Transformer \cite{dong_icassp2018} & Label smoothing \cite{szegedy_cvpr2016}& 7.1 & 5.1 & 5.3 \\ 
      Transformer \cite{dong_icassp2018} & KD \cite{bai_interspeech2019} & 7.1 & 5.2 & 5.4 \\ 
      Hierarchical transformer & - & 7.0 & 5.3 & 5.5 \\ 
      Hierarchical transformer & Label smoothing \cite{szegedy_cvpr2016} & 6.7 & 4.5 & 4.8 \\      
      Hierarchical transformer & Large-context KD & {\bf 6.5} & {\bf 4.3} & {\bf 4.5} \\\hline
    \end{tabular}
  \end{center}
  \vspace{-6mm}
\end{table}

\section{Conclusions}
We proposed a hierarchical transformer-based large-context E2E-ASR model and a large-context knowledge distillation method as an effective training method. The key advantage of the proposed model is that long-range sequential contexts beyond utterance boundaries can be captured while retaining the strengths of transformer-based E2E-ASR. Our large-context knowledge distillation method enables a large-context E2E-ASR model to use long-range contexts by mimicking the behavior of a large-context language model. Experimental results on discourse ASR tasks indicate that the proposed model and proposed training method effectively improves ASR performance.

\footnotesize{

}

\end{document}